%% file: acl_latex.tex
\pdfoutput=1

\documentclass[11pt]{article}

\usepackage{acl}

\usepackage{multirow}
\usepackage{booktabs,arydshln}
\usepackage{amsmath}
\newcommand{\ra}[1]{\renewcommand{\arraystretch}{#1}}
\makeatletter
\def\adl@drawiv#1#2#3{%
        \hskip.5\tabcolsep
        \xleaders#3{#2.5\@tempdimb #1{1}#2.5\@tempdimb}%
                #2\z@ plus1fil minus1fil\relax
        \hskip.5\tabcolsep}
\newcommand{\cdashlinelr}[1]{%
  \noalign{\vskip\aboverulesep
           \global\let\@dashdrawstore\adl@draw
           \global\let\adl@draw\adl@drawiv}
  \cdashline{#1}
  \noalign{\global\let\adl@draw\@dashdrawstore
           \vskip\belowrulesep}}
\makeatother

\usepackage{times}
\usepackage{latexsym}

\usepackage[T1]{fontenc}

\usepackage[utf8]{inputenc}

\usepackage{microtype}

\usepackage{inconsolata}

\usepackage[subtle]{savetrees}

\title{Stance Detection on Social Media with Fine-Tuned Large Language Models}

\author{
    Ilker Gül \\
    EPFL \\
    Lausanne, Switzerland \\
    \texttt{ilker.gul@epfl.ch} \\\And
    Rémi Lebret \\
    EPFL \\
    Lausanne, Switzerland \\
    \texttt{remi.lebret@epfl.ch} \\\And
    Karl Aberer \\
    EPFL \\
    Lausanne, Switzerland \\
    \texttt{karl.aberer@epfl.ch} \\
}

\begin{document}
\maketitle
\begin{abstract}

\input{abstract}

\end{abstract}

\section{Introduction}

\input{introduction}


\section{Methods}
\input{methods}


\section{Results}
\input{results}


\section{Discussion}
\input{conclusion}

\section*{Limitations}

\input{limitations}

\section*{Ethical Considerations}

\input{ethical_considerations}
\bibliography{anthology,custom}

\appendix

\section{Prompting Technique}
\label{sec:appendix2}
\input{app_prompt}

\end{document}

%% file: abstract.tex
Stance detection, a key task in natural language processing, determines an author's viewpoint based on textual analysis. This study evaluates the evolution of stance detection methods, transitioning from early machine learning approaches to the groundbreaking BERT model, and eventually to modern Large Language Models (LLMs) such as ChatGPT, LLaMa-2, and Mistral-7B. While ChatGPT's closed-source nature and associated costs present challenges, the open-source models like LLaMa-2 and Mistral-7B offers an encouraging alternative. Initially, our research focused on fine-tuning ChatGPT, LLaMa-2, and Mistral-7B using several publicly available datasets. Subsequently, to provide a comprehensive comparison, we assess the performance of these models in zero-shot and few-shot learning scenarios. The results underscore the exceptional ability of LLMs in accurately detecting stance, with all tested models surpassing existing benchmarks. Notably, LLaMa-2 and Mistral-7B demonstrate remarkable efficiency and potential for stance detection, despite their smaller sizes compared to ChatGPT. This study emphasizes the potential of LLMs in stance detection and calls for more extensive research in this field. 


%% file: introduction.tex
Stance detection plays a crucial role in analyzing social media content, aiming to identify an author's viewpoint—supportive, oppositional, or neutral—towards various subjects, from political figures to environmental policies. This process, relying on nuanced textual analysis across platforms like X (formerly Twitter), presents both opportunities and challenges for accurate interpretation \cite{hasan-ng-2013-stance, 10.1145/3369026, DBLP:journals/corr/abs-2006-03644}, key to gaining insights into public sentiment and societal trends. These insights are invaluable for research in data extraction, policy analysis, and beyond \cite{ siddiqua-etal-2019-tweet, darwish2017trump, glandt-etal-2021-stance}. As natural language processing (NLP) and social computing fields advance, they provide enhanced tools for effectively extracting and analyzing stances from social media texts, reflecting the evolving landscape of online discourse.

Initially, stance detection predominantly utilized rule-based and traditional machine learning approaches, with support vector machines (SVM) marking a significant early benchmark \cite{anand-etal-2011-cats, walker2012your, mohammad-etal-2016-semeval}. The introduction of deep learning models marked a pivotal shift, improving over traditional methods due to their ability to process large datasets and uncover complex patterns \cite{wei-etal-2016-pkudblab, zarrella-marsh-2016-mitre, dey2018topical, huang2018aspect, zhang-etal-2019-aspect}. The significant advancements continued with the development of pre-trained language models (PLMs) such as BERT, which facilitated a deeper understanding of textual context \cite{devlin-etal-2019-bert, li-etal-2021-p, kawintiranon-singh-2021-knowledge}.

The evolution of Large Language Models (LLMs) like OpenAI's ChatGPT and Meta AI's LLaMa-2 represents a major leap in NLP \cite{qin2023chatgpt, touvron2023llama}. These models, trained on vast datasets, are adept at mimicking the subtleties of human language with remarkable accuracy \cite{zhao2023survey, yin-etal-2023-large}. Both open-source and adaptable, LLaMa-2 and Mistral-7B enhance NLP technology accessibility and facilitate research without proprietary constraints \cite{jiang2023mistral}. Unlike their predecessors, LLMs use flexible prompting strategies, allowing them to perform a wide array of NLP tasks directly, including stance detection. This capability to navigate the intricate relationship between textual context and target subjects, accurately aligning with the author's intended stance, underscores their advanced understanding of language \cite{brown2020language, chowdhery2022palm, wei2023larger}.

The implementation of prompting strategies represents a significant departure from traditional NLP model training methods. By employing these strategies, LLMs can generate predictions without the extensive fine-tuning typically required, showcasing their versatility across various tasks. This methodological shift has not only simplified the application of LLMs but also expanded their utility, making them particularly effective for tasks that require a deep understanding of language nuances, such as stance detection \cite{bang-etal-2023-multitask, ouyang2022training}.

Fine-tuning tailors LLMs like ChatGPT, LLaMa-2, and Mistral-7B to specific tasks, significantly enhancing their precision and relevance for contextually aware stance detection on social media platforms \cite{zhang2023llamaadapter}. This adaptation to the unique language and style of social media discourse enables these models to outperform traditional methods, demonstrating superior performance in identifying sentiments and viewpoints.

The flexibility of LLMs in social media analysis is highlighted by their use of advanced techniques such as fine-tuning, chain of thought prompting \cite{chen2023you}, and both zero-shot and few-shot learning \cite{zhang2023stance, cruickshank2023use, aiyappa-etal-2023-trust}. These methods help navigate complex online discourse, enhancing the models' ability to interpret both explicit and implicit content effectively \cite{gatto2023chainofthought, lan2023stance}.

Our study evaluates the enhanced performance of fine-tuned LLMs using Twitter datasets, now known as X, to analyze a wide range of user opinions. We aim to show that fine-tuning significantly improves model understanding of user viewpoints, offering a deeper insight into online discourse. This research underscores the advantages of fine-tuning in NLP, particularly in stance detection, highlighting its superiority over traditional and less-tailored methods.


%% file: methods.tex
\subsection{Datasets and Evaluation Metrics}

\sloppy
\textbf{Datasets.} To assess the performance of our fine-tuned LLMs, we employed three publicly available datasets. The SemEval-2016 Dataset~\cite{mohammad-etal-2016-semeval} includes diverse societal and political topics, with stances categorized as Favor, Against, or None. This dataset examines a range of targets including political figures such as Donald Trump and Hillary Clinton, as well as broad issues like the Feminist Movement and Climate Change. It offers a total of 2,914 training instances and 1,956 testing instances. Detailed breakdowns of the training and testing data for each target are provided in Table~\ref{tab:dataset-distribution}, underscoring the dataset's utility for analyzing varied viewpoints.

\begin{table}[ht]
\centering
\begin{tabular}{lcc}
\hline
\textbf{Target} & \textbf{\# train} & \textbf{\# test} \\ \hline
Atheism                 & 513    & 220   \\
Climate Change is Concern & 395    & 169   \\
Feminist Movement       & 664    & 285   \\
Hillary Clinton         & 689    & 295   \\
Legalization of Abortion & 653    & 280   \\
Donald Trump             & 0    & 707   \\ \hline
\textbf{Total}          & 2914   & 1956  \\ \hline
\end{tabular}
\caption{Distribution of SemEval-2016 Dataset}
\label{tab:dataset-distribution}
\end{table}

The P-Stance Dataset~\cite{li-etal-2021-p}, on the other hand, narrows its focus to the political domain, featuring 21,574 labeled tweets that express either favor or opposition towards political figures Bernie Sanders, Donald Trump, and Joe Biden. Detailed statistics of this structured dataset are presented in Table~\ref{tab:p-stance-distribution}, offering a structured dataset for nuanced model training and analysis.

\begin{table}[ht]
\centering
\begin{tabular}{lccc}
\hline
\textbf{Target} & \textbf{\# train} & \textbf{\# validation} & \textbf{\# test} \\ \hline
Bernie Sanders   & 5,056  & 634   & 635   \\
Joe Biden        & 5,806  & 745   & 745   \\
Donald Trump     & 6,362  & 795   & 796   \\ \hline
\textbf{Total}   & 17,224 & 2,174 & 2,176 \\ \hline
\end{tabular}
\caption{Distribution of the P-Stance Dataset}
\label{tab:p-stance-distribution}
\end{table}

The Twitter Stance Election 2020 dataset, presented in Table~\ref{tab:twitter-stance2020-distribution} is derived from a larger corpus of over 5 million tweets, collected between January and September 2020 using election-related hashtags, to capture a comprehensive view of public opinion \cite{kawintiranon-singh-2021-knowledge}. 

\begin{table}[ht]
\centering
\begin{tabular}{lccc}
\hline
\textbf{Candidate} & \textbf{\# Train} & \textbf{\# Test} \\ \hline
Joe Biden          & 875  & 375 \\
Donald Trump       & 875  & 375  \\ \hline
\textbf{Total}     & 1,750 & 750  \\ \hline
\end{tabular}
\caption{Distribution of the Twitter Stance 2020 Dataset}
\label{tab:twitter-stance2020-distribution}
\end{table}

This dataset is labeled through Amazon Mechanical Turk \cite{crowston2012amazon} to ensure a balanced representation, it serves as a pivotal resource for analyzing the public's stance towards the candidates in the 2020 US Presidential election.


\textbf{Evaluation Metrics.} In line with the standards set by previous studies \cite{mohammad-etal-2016-semeval, mohammad2017stance}, we adopt \( F_{\mathrm{avg}} \) as our primary evaluation metric, enabling us to assess their precision, recall, and overall effectiveness in stance detection across varied contexts. This metric, \( F_{\mathrm{avg}} \), computes the average of the \( F1 \) scores for the 'favor' and 'against' classes.
\[
F_{\text{avg}} = \frac{F_{\text{favor}} + F_{\text{against}}}{2}
\]

For the analysis of the twitterstance2020, we adhere to the methodology outlined by the dataset's original authors \cite{kawintiranon-singh-2021-knowledge}, employing the F1-macro score across the three classes. This approach ensures consistency with the established evaluation framework and facilitates direct comparison with previous findings regarding this dataset.


\subsection{Models}


In our experiments, we utilized the \texttt{ChatGPT-1106} (gpt-3.5-turbo-1106) for fine-tuning, zero-shot, and few-shot tasks, as per OpenAI's guidelines\footnote{\href{https://platform.openai.com/docs/guides/fine-tuning}{https://platform.openai.com/docs/guides/fine-tuning}}. We fine-tuned the model over three epochs, creating the \texttt{ChatGPT-ft} variant. This choice was governed by the closed-source nature and cost considerations of ChatGPT's API.

We also explored open-source alternatives, \texttt{LLaMa-2} and \texttt{Mistral-7B}, developed respectively by Meta AI and Mistral AI. For \texttt{LLaMa-2} models (\texttt{LLaMa-2-7b} and \texttt{LLaMa-2-13b}), and the \texttt{Mistral-7b}, fine-tuning was adapted to dataset specifics: three epochs for SemEval-2016 and twitterstance2020, and one epoch for the larger P-Stance dataset to mitigate overfitting.

Post fine-tuning, the models were designated as \texttt{LLaMa-2-7b-ft}, \texttt{LLaMa-2-13b-ft}, and \texttt{Mistral-7b-ft}. We employed Low-Rank Adaptation (LoRA) \cite{hu2021lora} to refine the self-attention queries and values, through the \texttt{Lit-GPT} framework \cite{litgpt-2023}. This approach was aimed at enhancing the stance detection capabilities of the models.

To detail the fine-tuning parameters, the selected hyperparameters for LoRA included a rank of 8, an \(\alpha\) of 16, and a dropout rate of 0.05. We adopted a warmup strategy using 10\% of the training data, spreading the training over three epochs, with a learning rate of \(3 \times 10^{-4}\) and a batch size of 128. The training was conducted on single NVIDIA A100 GPUs with 40GB of RAM, using bfloat16 precision.

The training sessions for \texttt{LLaMa-2-7b-ft} and \texttt{Mistral-7b-ft} on the SemEval-2016 dataset were completed in approximately 20 minutes, with \texttt{LLaMa-2-13b-ft} requiring about 30 minutes. This detail ensures the reproducibility of our results and the assessment of our method's robustness.

We tested the impact of training data volumes (70\%, 50\%, and 30\% of the total dataset) on model performance, identifying the optimal training volume that balances accuracy with resource efficiency, highlighting the effectiveness of fine-tuning in stance detection.

Our study utilized both the base versions of each model for fine-tuning and their instruction-tuned variants for the zero-shot and few-shot prompting evaluations. The instruction-tuned versions, \texttt{LLaMa-2-7b-chat}, \texttt{LLaMa-2-13b-chat}, and \texttt{Mistral-7b-Instruct}, were enhanced by pre-training on instruction datasets with over 1 million human annotations, improving their ability to follow complex instructions \cite{touvron2023llama}.

For our analysis, we conducted zero-shot and few-shot experiments with \texttt{ChatGPT} and the instruction-tuned models. In few-shot prompting tests, we utilized three training samples from different stance categories, selected for their relevance to our test cases using the ``all-MiniLM-L6-v2'' model from the sentence-transformers library \footnote{\href{https://huggingface.co/sentence-transformers}{https://huggingface.co/sentence-transformers}}. This method provided insights into each model's ability to generalize from minimal examples and revealed their performance across various learning settings.

\subsection{Prompting Details}

For the ChatGPT model, we employed specific prompting methods for each dataset following the guideline provided by OpenAI\footnote{\href{https://platform.openai.com/docs/guides/prompt-engineering}{https://platform.openai.com/docs/guides/prompt-engineering}}. 
For the LLaMa-2 model, our prompting strategy was inspired by the template samples available in HuggingFace's resources\footnote{\href{https://huggingface.co/blog/llama2}{https://huggingface.co/blog/llama2}}. 
For the Mistral model, we devised our prompting approach based on best practices and recommendations detailed in their website\footnote{\href{https://docs.mistral.ai/guides/prompting-capabilities/}{https://docs.mistral.ai/guides/prompting-capabilities/}}.
Detailed specifications of the prompts used for each dataset can be found in the Appendix~\ref{sec:appendix2}.

\subsection{Baselines}
We have selected various stance detection models as our baselines, with different foundational architectures and approaches. We include BiLSTM and BiCond models, which utilize bidirectional LSTM layers to capture language context from both directions, crucial for nuanced stance detection~\cite{augenstein-etal-2016-stance}. MemNet~\cite{tang-etal-2016-aspect}, from memory networks, retains key sentiment aspects from text, enhancing context-aware analysis.

AoA~\cite{huang2018aspect} and TAN~\cite{du2017stance} incorporate attention mechanisms; AoA uses hierarchical attention to prioritize different text parts, while TAN applies target-specific attention to focus on the target-text relation. ASGCN \cite{zhang2019aspect} uses graph neural network techniques to understand textual interdependencies effectively. AT-JSS-Lex \cite{li-caragea-2019-multi} integrates sentiment analysis with stance detection, utilizing a sentiment lexicon to leverage sentiment signals for better stance prediction. TPDG \cite{liang2021target} prioritizes target-dependent representations in its detection approach.

StSQA \cite{zhang2023investigating} introduces a unique method of stance detection by teaching models like ChatGPT to perform stance detection with minimal examples, exemplified through a 1-shot learning scenario. TR-Tweet+COT \cite{gatto2023chainofthought} introduces COT Embeddings to enhance Chain-of-Thought (COT) prompting in LLMs for social media stance detection, integrating them with RoBERTa to handle implicit stances. 

Lastly, the COLA framework \cite{lan2023stance} employs role-specific LLMs to dissect and deduce stances from text, addressing the multifaceted nature of stance detection and culminating in an agent that synthesizes these insights for accurate stance detection.

%% file: results.tex
\subsection{Controversial Topics on Twitter}

\subsubsection{Fine-Tuned Models vs Baselines}

\begin{table}[!htb]
\centering
\footnotesize
\setlength{\tabcolsep}{4pt} 
\begin{tabular}{@{}lcccccc@{}}
\toprule
 & \textbf{FM} & \textbf{HC} & \textbf{LA} & \textbf{A} & \textbf{CC} & \textbf{DT}  \\
\midrule
\textbf{Baseline models}\\
\addlinespace[4pt]
BiCond & 61.4 & 59.8 & 54.5 & - & - & 59.0 \\
\addlinespace[4pt]
MemNet & 57.8 & 60.3 & 61.0 & - & - & -\\
\addlinespace[4pt]
AoA & 60.0 & 58.2 & 62.4 & - & - & - \\
\addlinespace[4pt]
TAN & 55.8 & 65.4 & 63.7 & 59.3 & 53.5 & - \\
\addlinespace[4pt]
ASGCN & 58.7 & 64.3 & 62.9 & - & - & 58.7\\
\addlinespace[4pt]
AT-JSS-Lex & 61.5 & 68.3 & 68.4 & 69.2 & 59.2 & - \\
\addlinespace[4pt]
TPDG & 67.3 & 73.4 & 74.7 & - & - & 63.0 \\
\addlinespace[4pt]
TR-Tweet+COT & 70.6 & 78.7 & 63.8 & 72.9 & 54.1 & - \\
\addlinespace[4pt]
COLA & 69.1 & 75.9 & 71.0 & 62.3 & 64.0 & 71.2 \\
\midrule
\textbf{Fine-tuned LLMs}\\
\addlinespace[4pt]
ChatGPT-ft & \textbf{79.7} & 83.4 & 72.6 & \textbf{81.3} & \textbf{86.2} & 70.4\\
\addlinespace[4pt]
LLaMa-2-7b-ft & 73.3 & 84.2 & 71.2 & 78.9 & 69.8 & \textbf{72.0}\\
\addlinespace[4pt]
LLaMa-2-13b-ft & 76.0 & \textbf{86.5} & 72.5 & 76.9 & 80.4 & 70.9\\
\addlinespace[4pt]
Mistral-7b-ft & 78.7 & 85.0 & \textbf{76.0} & 74.7 & 71.8 & 68.6\\
\bottomrule
\end{tabular}
\caption{SemEval-2016 Dataset performance with \( F_{avg} \). \textit{Note: FM - Feminist Movement, HC - Hillary Clinton, LA - Legalization of Abortion, Ath - Atheism, CC - Climate Change is a Real Concern, DT - Donald Trump. Each acronym represents a specific target within the dataset, assessed with \( F_{avg} \) scores.}}
\label{semeval2016-results}
\end{table}

The analysis of the SemEval-2016 Dataset, presented in Table \ref{semeval2016-results}, reveals significant findings regarding the performance enhancements achieved through fine-tuning of language models on specific topics such as the Feminist Movement (FM), Hillary Clinton (HC), and the Legalization of Abortion (LA). These topics were selected for detailed analysis due to the completeness and availability of comparative data.

Among the baseline models, TR-Tweet+COT and COLA demonstrated the power of leveraging advanced language model architectures, setting new benchmarks in stance detection with notable \(F_{avg}\) scores. For instance, TR-Tweet+COT achieved an \(F_{avg}\) of 70.6 for FM and COLA scored 75.9 for HC, illustrating significant improvements over traditional models. These advancements suggest that the deeper contextual understanding provided by LLM-based models is crucial for effectively navigating the complexities inherent in stance detection tasks.

The superior performance of fine-tuned models such as \texttt{ChatGPT-ft}, which achieved an \(F_{avg}\) score of 79.7 for FM, reflects the effectiveness of fine-tuning in enhancing model sensitivity to nuanced discussions. This model outperformed the top baseline model by over 9 points, highlighting the transformative impact of fine-tuning on model capabilities.

In the case of political figures like Hillary Clinton, \texttt{LLaMa-2-13b-ft} recorded the highest \(F_{avg}\) of 86.5, surpassing all baseline models. This performance underscores the potential of fine-tuned models to accurately identify and analyze sentiments, benefiting from the refined model parameters tailored specifically to the task.

For sensitive topics such as the Legalization of Abortion, \texttt{Mistral-7b-ft} demonstrated substantial performance improvement, achieving an \(F_{avg}\) of 76.0, which indicates the model's enhanced ability to handle complex discourse following fine-tuning. This confirms the adaptability of LLMs through fine-tuning, which enables them to better capture and interpret the subtleties within the data crucial for accurate stance detection.


\subsubsection{Fine-Tuning Impact: Comparing Training Scales}

\begin{table}[!htb]
\centering
\footnotesize 
\setlength{\tabcolsep}{4pt} 
\renewcommand{\arraystretch}{1.4} 
\begin{tabular}{@{}l@{\hspace{6pt}}c@{\hspace{6pt}}c@{\hspace{5pt}}c@{\hspace{5pt}}c@{\hspace{5pt}}c@{}}
\toprule
\textbf{Models} & & & \textbf{FM} & \textbf{HC} & \textbf{LA} \\
\midrule
 & & & \multicolumn{3}{c}{\textit{Zero-shot Prompting}} \\
 \cmidrule{4-6}
ChatGPT-1106 & & & 74.6 & 82.8 & 59.6 \\
LLaMa-2-7b-chat & & & 51.6 & 63.9 & 49.2 \\
LLaMa-2-13b-chat & & &  55.0 & 61.5 & 45.9 \\
Mistral-7b-instruct & & & 50.7  & 66.3  & 50.1  \\
\midrule
 & & & \multicolumn{3}{c}{\textit{Few-shot Prompting}} \\
  \cmidrule{4-6}
ChatGPT-1106  & & & 75.6 & 82.9 & 62.4  \\
LLaMa-2-7b-chat & & & 62.1 & 72.7 & 49.1 \\
LLaMa-2-13b-chat & & & 62.6 & 75.4 & 51.1 \\
Mistral-7b-instruct & & & 59.7 & 76.6 & 51.2 \\
\midrule
 & \multicolumn{2}{r}{\textit{Train Size}} & \multicolumn{3}{c}{\textit{Fine-Tuning}} \\
 \cmidrule{2-6}
ChatGPT-ft & 100\% & &  79.7 & 83.4 & 72.6 \\
\cdashlinelr{1-6}
\multirow{4}{*}{LLaMa-2-7b-ft} &  100\% & & 73.3 & 84.2 & 71.2 \\
 & 70\% & &  71.5 & 84.4 & 71.1 \\
 & 50\% & &  70.7 & 83.2 & 67.4 \\
 & 30\% & &  67.7 & 82.4 & 58.1 \\
\cdashlinelr{1-6}
\multirow{4}{*}{LLaMa-2-13b-ft} &  100\% & & 76.0 & 86.5 & 72.5 \\
 & 70\% & &  76.8 & 86.3 & 71.2 \\
 & 50\% & &  68.3 & 86.1 & 68.3 \\
 & 30\% & &  63.7 & 84.8 & 64.1 \\
\cdashlinelr{1-6}
\multirow{4}{*}{Mistral-7b-ft} &  100\% & & 78.7 & 85.0 & 76.0 \\
 & 70\% & &  76.3 & 80.8 & 69.9 \\
 & 50\% & &  71.7 & 84.8 & 72.8 \\
 & 30\% & &  73.8 & 83.1 & 73.7 \\
\bottomrule
\end{tabular}
\caption{\label{complete-comparison-results} \( F_{\mathrm{avg}} \) scores among different configurations for SemEval-2016 Dataset}
\end{table}

The results presented in Table \ref{complete-comparison-results} demonstrate the impact of different training strategies on the performance of state-of-the-art language models across the SemEval-2016 dataset. This table compares zero-shot, few-shot, and fine-tuned performance, emphasizing the substantial gains achieved through fine-tuning.

ChatGPT, when fine-tuned, exhibited significant improvements in \(F_{\mathrm{avg}}\) scores, highlighting the effectiveness of adapting pre-trained models to specific datasets. For instance, the fine-tuned ChatGPT model achieved an \(F_{\mathrm{avg}}\) score of 79.7 for the Feminist Movement (FM), which was a 5.1 point increase over its zero-shot performance. This substantial improvement underscores the importance of fine-tuning in extracting more precise and contextually relevant information from complex datasets.

LLaMa-2 models, particularly the 7b variant, also demonstrated notable performance enhancements. The zero-shot configuration scored 51.6 for FM, which jumped to 73.3 upon fine-tuning—an increase that highlights the model’s capacity to significantly refine its stance detection capabilities with comprehensive training.

Analysis of the effects of reduced training data volumes on model performance revealed a nuanced pattern: while reductions typically led to lower \(F_{\mathrm{avg}}\) scores, the decrease was not always proportional, indicating efficient learning even from reduced datasets. For example, the LLaMa-2-7b model trained with 70\% of the data managed nearly equivalent performance to the fully trained model in the HC category (84.4 vs. 84.2), suggesting that a substantial part of the model's learning had been achieved with the reduced data set.

The substantial performance gains in the fine-tuned models as compared to the zero-shot and few-shot configurations illustrate the critical role of fine-tuning in aligning the model's parameters with the intricacies of specific stance detection tasks. This is particularly evident in complex or controversial topics such as Legalization of Abortion (LA), where contextual understanding is crucial for accurate stance detection.

These results not only validate the effectiveness of fine-tuning in enhancing the performance of language models in stance detection but also highlight the potential efficiencies that can be realized even with reduced training volumes. This suggests a promising direction for future research where less data might be used more strategically to achieve high performance, especially in resource-constrained settings.

\subsection{Political Figures on Twitter}

\subsubsection{Fine-Tuned Models vs Baselines}
\label{sec:fine-tuned-vs-baselines}

Our study evaluates the efficacy of fine-tuned Large Language Models (LLMs) against baseline models in capturing nuanced public stances towards political figures, utilizing two distinct datasets: the P-Stance Dataset and the Twitter Stance Election Dataset 2020. We begin with an in-depth exploration of the P-Stance Dataset.


\begin{table}[!htb]
\centering
\ra{1.1}
\begin{tabular}{@{}lccc@{}}
\toprule
 & \textbf{Bernie} & \textbf{Biden} & \textbf{Trump} \\
\midrule
\textbf{Baseline Models}\\
BiLSTM & 63.9 & 69.5 & 72.0 \\
BiCond & 64.6 & 69.4 & 73.0 \\
MemNet & 72.8 & 77.6 & 77.7 \\
TAN & 72.0 & 77.9 & 77.5 \\
AoA & 71.7 & 77.8 & 77.7 \\
ASGCN & 70.8 & 78.4 & 77.0 \\
StSQA & 80.8 & 82.6 & 85.7 \\
\midrule
\textbf{Fine-tuned LLMs}\\
ChatGPT-ft & \textbf{81.8} & \textbf{89.7} & \textbf{91.9} \\
LLaMa-2-7b-ft & 79.0 & 87.2 & 89.8 \\
LLaMa-2-13b-ft & 81.0 & 89.0 & 88.9 \\
Mistral-7b-ft & 81.0 & 88.2 & 90.3 \\
\bottomrule
\end{tabular}
\caption{\label{pstance2021-results}
P-Stance Dataset performance  with \( F_{avg} \) scores
}
\end{table}

The results in Table~\ref{pstance2021-results} underscore the substantial impact of fine-tuning on stance detection accuracy. Notably, \texttt{ChatGPT-ft} demonstrates exceptional performance, significantly surpassing both baseline models and other fine-tuned counterparts. This indicates that the fine-tuning process, leveraging ChatGPT's extensive pre-trained knowledge, considerably enhances its ability to discern complex stances across various political figures.

A critical observation from the P-Stance Dataset is the pronounced performance differential between models, particularly in how fine-tuned LLMs like \texttt{ChatGPT-ft} and \texttt{Mistral-7b-ft} navigate complex stance-related nuances compared to more traditional models like \texttt{StSQA}. While \texttt{StSQA} performs robustly, the advanced adaptation strategies employed in LLMs ensure a deeper alignment with the nuanced demands of social discourse, enabling them to outperform traditional models significantly.

Next, we extend our investigation to the Twitter Stance Election Dataset 2020, focusing on the politically charged atmosphere of the 2020 U.S. presidential election. This dataset provides a rich context for further evaluating the refined capabilities of fine-tuned LLMs.

\begin{table}[!htb]
\centering
\ra{1.1}
\begin{tabular}{@{}lcc@{}}
\toprule
           & \multicolumn{1}{c}{\textbf{Biden}} & \multicolumn{1}{c}{\textbf{Trump}} \\ 
\midrule
\textbf{Baseline Models}\\
a-BERT          & 75.2                               & 77.2                               \\
SKEP            & 74.8                               & 77.2                               \\
KE-MLM-         & 74.3                               & 76.3                               \\
KE-MLM          & 75.7                               & 78.7                               \\ 
\midrule
\textbf{Fine-tuned LLMs}\\
ChatGPT-ft      & \textbf{85.1}                      & \textbf{85.6}                               \\
LLaMa-2-7b-ft   & 73.4                               & 74.1                               \\
LLaMa-2-13b-ft  & 74.1                               & 74.8                               \\
Mistral-7b-ft   & 74.1                               & 74.3                               \\ 
\bottomrule
\end{tabular}
\caption{F1-macro scores for Twitter Stance Election 2020 Dataset}
\label{tab:model_comparison}
\end{table}

The results from Table \ref{tab:model_comparison} reveal significant performance differences between the fine-tuned models and the baselines. Notably, \texttt{ChatGPT-ft} excels, achieving F1-macro scores of 85.1 for Biden and 85.6 for Trump, outstripping both baseline models like a-BERT and SKEP, and other fine-tuned LLMs such as \texttt{LLaMa-2} variants and \texttt{Mistral-7b-ft}. This superior performance may be attributed to \texttt{ChatGPT-ft} having been extensively pretrained on a diverse dataset that likely includes significant coverage of political figures. This extensive pre-training enables the model to enhance its performance in tasks involving political context.

The \texttt{LLaMa-2} models and \texttt{Mistral-7b-ft} demonstrate notable improvements over baseline models but still do not reach the impressive performance levels of \texttt{ChatGPT-ft}. This discrepancy highlights the significant influence of model architecture and the nuances of pre-training on the effectiveness of subsequent fine-tuning efforts. Despite undergoing fine-tuning under comparable conditions, the inherent architectural differences between models significantly affect their ability to predict accurate stances.

Given the multifaceted nature of social media content, the variation in performance among models like \texttt{ChatGPT-ft}, \texttt{LLaMa-2}, and \texttt{Mistral-7b-ft} in stance detection tasks is particularly revealing. It accentuates the necessity for model-specific fine-tuning strategies that cater to the unique strengths and limitations of each architecture. This approach underscores the need for continuous optimization to keep pace with the rapidly evolving landscape of social media, particularly in the realm of political content, which is consistently dynamic and continuously evolving.

\subsubsection{Fine-Tuning Impact: Comparing Training Scales}

Building on our initial findings in Section~\ref{sec:fine-tuned-vs-baselines}, we now examine the nuanced impact of fine-tuning on model performance, particularly focusing on the P-Stance Dataset due to its larger size compared to the Twitter Stance Election Dataset 2020. This dataset allows for a detailed exploration of various training strategies, including zero-shot, few-shot, and fine-tuning across different scales of data.

\begin{table}[!htb]
\centering
\footnotesize
\setlength{\tabcolsep}{4pt}
\renewcommand{\arraystretch}{1.4}
\begin{tabular}{@{}l@{\hspace{6pt}}c@{\hspace{6pt}}c@{\hspace{5pt}}c@{\hspace{5pt}}c@{\hspace{5pt}}c@{}}
\toprule
\textbf{Models} & & & \textbf{Bernie} & \textbf{Biden} & \textbf{Trump} \\
\midrule
 & & & \multicolumn{3}{c}{\textit{Zero-shot Prompting}} \\
 \cmidrule{4-6}
ChatGPT-1106 & & & 75.2 & 82.6 & 73.7 \\
LLaMa2-7b-chat & & & 48.3 & 52.9 & 43.6 \\
LLaMa2-13b-chat & & & 49.8 & 53.7 & 45.3 \\
Mistral-7b-instruct & & & 43.6  & 48.7  & 40.2 \\
\midrule
& & & \multicolumn{3}{c}{\textit{Few-shot Prompting}} \\
\cmidrule{4-6}
ChatGPT-1106  & & & 78.9  & 86.4  & 84.3  \\
LLaMa2-7b-chat & & & 61.8 & 69.7 & 70.1 \\
LLaMa2-13b-chat & & & 60.7 & 70.5 & 73.2 \\
Mistral-7b-instruct & & & 60.2 & 68.3 & 76.8 \\
\midrule
 & \multicolumn{2}{r}{\textit{Train Size}} & \multicolumn{3}{c}{\textit{Fine-Tuning}} \\
 \cmidrule{2-6}
ChatGPT-ft & & 100\% & 81.8 & 89.7 & 91.9 \\
\cdashlinelr{1-6}
\multirow{4}{*}{LLaMa2-7b-ft} & & 100\% & 79.0 & 87.2 & 89.8 \\
& & 70\% & 81.7 & 86.9 & 89.4 \\
 & & 50\% & 80.8 & 87.5 & 89.7 \\
 & & 30\% & 79.7 & 87.0 & 88.3 \\
\cdashlinelr{1-6}
\multirow{4}{*}{LLaMa2-13b-ft} & & 100\% & 81.0 & 89.0 & 88.9 \\
 & & 70\% & 81.9 & 88.1 & 91.0 \\
 & & 50\% & 81.9 & 88.9 & 90.3 \\
& & 30\% & 80.4 & 88.6 & 90.1 \\
\cdashlinelr{1-6}
\multirow{4}{*}{Mistral-7b-ft} & & 100\% & 81.0 & 88.2 & 90.3 \\
& & 70\% & 80.1 & 87.8 & 89.3 \\
 & & 50\% & 80.9 & 87.3 & 88.0 \\
 & & 30\% & 80.3 & 86.9 & 87.2 \\
\bottomrule
\end{tabular}
\caption{P-Stance Dataset performance comparison (using \( F_{avg} \) scores)}
\label{pstance2021-results2}
\end{table}

Table~\ref{pstance2021-results2} presents a comprehensive evaluation of different models' performances on the P-Stance Dataset across three settings: zero-shot, few-shot, and fine-tuning. This analysis particularly highlights the strengths and limitations of each training strategy.

\texttt{ChatGPT-1106} stands out in the zero-shot setting, particularly for its handling of content related to Joe Biden, indicating strong generalization capabilities possibly due to extensive pretraining that included relevant political topics. Its fine-tuned version, \texttt{ChatGPT-ft}, not only sustains but significantly enhances this performance, achieving the highest \( F_{avg} \) scores across all models tested. This suggests that fine-tuning, tailored to specific stance detection tasks, sharply increases the model’s precision and relevance.

The \texttt{LLaMa-2} models in their zero-shot variants initially struggle with stance detection but see substantial improvements after fine-tuning. The \texttt{LLaMa-2-13b-ft} demonstrates a slight advantage over the \texttt{LLaMa-2-7b-ft}, particularly in handling complex linguistic data, which can be attributed to its larger capacity and possibly more nuanced understanding of context. Few-shot learning, while improving over zero-shot results, does not achieve the high accuracy seen with fully fine-tuned models, yet it still offers valuable gains that can be crucial when extensive training data is not available.

\texttt{Mistral-7b-instruct} shows that even moderate zero-shot performance can be significantly elevated through fine-tuning, particularly highlighting its enhanced performance against Donald Trump. This underscores the potential of fine-tuning to adapt a model to the nuanced requirements of specific datasets or domains, enhancing its operational effectiveness.

Across all models, fine-tuning clearly emerges as the most effective strategy for optimizing performance in stance detection, significantly outperforming the more limited zero-shot and few-shot approaches. This analysis not only confirms the value of comprehensive fine-tuning but also illustrates how different scales of training data impact model efficiency. For instance, both \texttt{LLaMa-2-7b-ft} and \texttt{LLaMa-2-13b-ft} models exhibit near-peak performance even when trained with only 70\% of the full dataset, suggesting that these models can achieve high efficiency and effectiveness without maximum data inputs. Conversely, \texttt{Mistral-7b-ft} maintains strong performance with reduced data, indicating its robustness and adaptability under constrained data conditions.

These findings highlight the importance of customizing fine-tuning strategies to each model's unique characteristics and the specific challenges of the task at hand. They suggest that while fine-tuning is critical for achieving optimal performance, the amount of data used for training can be adjusted without necessarily compromising the outcomes.

%% file: conclusion.tex
Our experiments with the SemEval-2016 and P-Stance 2021 datasets demonstrate the efficacy of fine-tuned Large Language Models (LLMs) in stance detection. Notably, the \texttt{ChatGPT-ft} model consistently outperformed others, as detailed in Tables \ref{semeval2016-results}, \ref{pstance2021-results}, and \ref{tab:model_comparison}. The \texttt{LLaMa-2} models and \texttt{Mistral-7b} also showed significant capabilities, underscoring the potential of LLMs in this nuanced task.

Our comparative analysis revealed that larger model size does not automatically equate to better performance. For instance, \texttt{LLaMa-2-13b-ft}, despite its larger size, did not consistently outperform the smaller \texttt{LLaMa-2-7b-ft} or \texttt{Mistral-7b-ft}. This finding highlights that factors like fine-tuning, dataset specifics, and architectural design are pivotal in achieving high performance. It suggests the importance of a balanced approach in model development that considers size, data handling, and design intricacies to optimize systems for stance detection.

Variations in model performance across different domains, such as the handling of topics like Hillary Clinton, underscore the models' sensitivity to domain-specific nuances. This sensitivity likely originates from the datasets used during initial pre-training, which may introduce biases or ingrained domain knowledge, impacting fine-tuned performance. Accurately interpreting stance across various topics is crucial for applications like content moderation, trend analysis, and misinformation detection in social media analysis.

Topics with less prevalent coverage in training data, like the Legalization of Abortion, showed significant improvements post-fine-tuning. This improvement points to the potential for tailored fine-tuning strategies to address gaps in model knowledge and enhance stance detection across a broader spectrum of subjects.

Overall, this analysis emphasizes the critical role of fine-tuning in enhancing stance detection performance. The comparable outcomes achieved by both \texttt{LLaMa-2} and \texttt{Mistral} models with less data suggest that more efficient model training strategies are possible, proposing that strategic data reduction can lead to highly effective models. It stresses the necessity of fine-tuning for achieving optimal performance and advocates for refined and strategic training approaches for specialized tasks.

In conclusion, fine-tuning not only optimizes LLMs for specific discourse contexts but also significantly outperforms traditional methodologies, showcasing the robustness and adaptability of fine-tuned LLMs in understanding diverse and complex societal issues. These results offer deep insights into stance detection, highlighting the robustness and adaptability of fine-tuned LLMs in handling the complexities of human communication on social media platforms.

\section{Conclusion}

In conclusion, our exploration of stance detection through the lenses of ChatGPT, LLaMa-2, and Mistral models demonstrates the significant advancements and potential these models bring to the field of NLP. With superior performance, these models are frontrunners in accurately understanding and interpreting textual data, notably outperforming traditional methods like zero-shot and few-shot prompting. This superiority is particularly crucial in social media analysis, where accurate detection of stance can inform a range of applications from sentiment analysis to misinformation detection.

Stance detection plays a pivotal role by providing deeper insights into public opinion, trends, and behaviors across a variety of topics. The capability of LLMs to process and analyze vast amounts of textual data makes them indispensable in navigating the complex landscapes of social media. Through fine-tuning, these models achieve a precision and adaptability essential for dissecting intricate expressions of stance, enhancing our ability to monitor, understand, and engage with societal discourse in meaningful ways.

Looking forward, the evolution of LLMs promises further advancements in stance detection and broader NLP applications. This progression marks a significant leap towards refining and improving models. As these models become more integrated into social media analysis tools, they will undoubtedly enrich our understanding of digital communication. 

However, it's important to address potential challenges and ethical concerns such as data privacy, model bias, and the environmental impact of training large models. Addressing these issues will be crucial as we anticipate profound implications for academia, industry, and beyond. This future, where technology and human communication converge, highlights the increasingly sophisticated and impactful ways we interact with and through digital platforms.


%% file: limitations.tex
In our research on stance detection using ChatGPT, we encountered several limitations that impacted our methodology and results. A primary constraint was ChatGPT's exclusivity, which restricts access to its features solely through its API. This limitation confined our ability to modify the model, with the number of fine-tuning epochs being the only adjustable hyperparameter.

The cost of training with ChatGPT, based on OpenAI's pricing structure of \$0.008 per 1,000 tokens, offers an affordable option compared to many other machine learning training expenses. For instance, fine-tuning a dataset with 100,000 tokens over three epochs costs approximately \$2.40 USD. Even training on larger datasets like SemEval-2016, which involves a more substantial number of tokens, is estimated at around \$21.77 USD. While the amount may seem manageable within typical research budgets, it's important to recognize that financial cost can still be a primary concern, particularly for smaller institutions or projects with limited funding. Alongside this, issues such as privacy concerns, latency, and the reliability of API calls also pose substantial challenges, each contributing significantly to the complexities of the research process.

We observed an average response time of about 2.50 seconds per API request, with frequent errors when processing large datasets iteratively. These issues underscore the potential benefits of hosting models on-premises for uninterrupted research operations. Moreover, privacy concerns emerge when handling proprietary or sensitive data, discouraging the use of external services for projects requiring strict confidentiality.

For the LLaMa-2 models, our fine-tuning efforts were constrained by computational resources. While the LLaMa-2 7b and 13b models could be fine-tuned using NVIDIA A100 GPUs with 40GB, the 70b model's requirements exceeded our available computing power. This highlights the importance of high-end resources for training more complex models.

Finally, training dataset size for specific targets within the SemEval-2016 dataset limited our stance detection capabilities, whereas the P-Stance dataset provided a more robust training resource. This suggests that larger datasets could enhance LLMs' performance in stance detection, particularly for social media analysis.

Efficient and cost-effective analysis of social media content using LLMs requires substantial computational resources and careful financial planning to accommodate the costs of training and fine-tuning on large datasets. These constraints necessitate a balanced approach to leveraging LLMs for social media analysis, weighing the benefits of advanced capabilities against the associated costs.

%% file: ethical_considerations.tex
In the course of this research, it's crucial to acknowledge the potential limitations of Large Language Models. Both ChatGPT and Llama 2, like other LLMs, may produce inaccurate information about targets present in stance detection datasets. Such inaccuracies can emerge from various factors inherent to algorithmic predictions and inherent model limitations.

This research relied on publicly available datasets for the fine-tuning of LLMs. The primary goal in using these datasets was academic research. At no stage was there an intention to produce or support biased predictions. 


%% file: app_prompt.tex
In our fine-tuning process, structured prompts were essential in creating the training and test datasets for the LLMs. The prompts are designed to offer context, guidelines, and the exact task the model is expected to accomplish. In this section, we provide a detailed overview of the prompts utilized for each dataset while fine-tuning ChatGPT.

\subsection{ChatGPT Fine-tuning Prompts}

\subsubsection{SemEval-2016 Template}

For the SemEval-2016 dataset, the following structured prompt was utilized:

\begin{quote}
\textbf{\#\#\# Instruction:} 

Analyze the tweet below in the following context: [topic]. Consider the text, subtext, regional and cultural references, and any implicit meanings to determine the stance expressed in the tweet towards the target. The possible stances are:
\begin{itemize}
    \item FAVOR: The tweet has a positive or supportive attitude towards the target, either explicitly or implicitly.
    \item AGAINST: The tweet opposes or criticizes the target, either explicitly or implicitly.
    \item NONE: The tweet is neutral or doesn't have a stance towards the target.
\end{itemize}
\textbf{Tweet:} [tweet]

\textbf{\#\#\# Question:} 

What is the stance expressed in the tweet towards the target "[target]"? 

Choose one of the following options: FAVOR, AGAINST, NONE.

\textbf{\#\#\# Answer:}
\end{quote}

For this prompt structure, placeholders are utilized: \texttt{[tweet]}, \texttt{[target]}, and \texttt{[topic]}. 

\begin{itemize}
    \item \textbf{\texttt{[tweet]}}: Represents the actual tweet being analyzed.
    
    \item \textbf{\texttt{[target]}}: Denotes what or whom the tweet's stance is directed at, whether directly or indirectly.
    
    \item \textbf{\texttt{[topic]}}: Offers a brief description of the \texttt{[target]}. Specifically for the SemEval-2016 dataset, this description was crafted by us to facilitate the understanding of the tweet's context.
\end{itemize}

When fine-tuning, these placeholders are substituted with real data, making it easier for the model to understand the context and identify the stance.

\subsubsection{P-Stance Template}

For the P-Stance dataset, the prompt tailored specifically for political domain analysis was:

\begin{quote}
\textbf{\#\#\# Instruction:} 

Analyze the following tweet, which is in the political domain, deeply. Consider any subtext, regional and cultural references, or implicit meanings to determine the tweet's stance towards the target. The possible stances are:
\begin{itemize}
    \item FAVOR: The tweet has a positive or supportive attitude towards the target, either explicitly or implicitly.
    \item AGAINST: The tweet opposes or criticizes the target, either explicitly or implicitly.
\end{itemize}
\textbf{Tweet:} [tweet]

\textbf{\#\#\# Question:} 

What is the stance of the tweet above towards the target "[target]"? 

Select from FAVOR or AGAINST.

\textbf{\#\#\# Answer:}
\end{quote}

The placeholders \texttt{[tweet]} and \texttt{[target]} are used in a similar manner as explained for the SemEval-2016 template above.

\subsubsection{Twitter Stance Election 2020 Template}

For the Twitter Stance Election 2020 dataset, the prompt tailored specifically for political domain analysis was:

\begin{quote}
\textbf{\#\#\# Instruction:} 

Analyze the following tweet, which is in the political domain, deeply. Consider any subtext, regional and cultural references, or implicit meanings to determine the tweet's stance towards the target. The possible stances are:
\begin{itemize}
    \item FAVOR: The tweet has a positive or supportive attitude towards the target, either explicitly or implicitly.
    \item AGAINST: The tweet opposes or criticizes the target, either explicitly or implicitly.
    \item NONE: The tweet is neutral or doesn't have a stance towards the target.
\end{itemize}
\textbf{Tweet:} [tweet]

\textbf{\#\#\# Question:} 

What is the stance of the tweet above towards the target "[target]"? 

Choose one of the following options: FAVOR, AGAINST, NONE.

\textbf{\#\#\# Answer:}
\end{quote}

The placeholders \texttt{[tweet]} and \texttt{[target]} are used in a similar manner as explained for the SemEval-2016 template above.

\subsection{Llama 2 Fine-Tuning Prompts}

\subsubsection{SemEval-2016 Llama 2 Template}

This prompt template focuses on detecting the stance in tweets using a structured instruction to guide the model:

\begin{quote}
[INST] \texttt{<<SYS>>} \\
You are a helpful, respectful, and honest assistant for stance detection for a given target. Always answer from the possible options given below as helpfully as possible. Stance detection is the process of determining whether the author of a tweet is in support of or against a given target. The target may not always be explicitly mentioned in the text, and the tweet's stance can be conveyed implicitly through subtext, regional and cultural references, or other implicit meanings. \\
The possible stances are:
\begin{itemize}
    \item support: The tweet has a positive or supportive attitude towards the target, either explicitly or implicitly.
    \item against: The tweet opposes or criticizes the target, either explicitly or implicitly.
    \item none: The tweet is neutral or doesn't have a stance towards the target.
\end{itemize}
\texttt{</SYS>} \\
Tweet: \texttt{[tweet]} \\
Stance towards the target [target]:[/INST]
\end{quote}

For this prompt structure, placeholders are utilized: \texttt{[tweet]} and \texttt{[target]}. 
\begin{itemize}
    \item \texttt{[tweet]}: Represents the actual tweet being analyzed.
    \item \texttt{[target]}: Denotes what or whom the tweet's stance is directed at.
\end{itemize}

\subsubsection{P-Stance LlaMa 2 Template}

This prompt template is specifically designed for analyzing tweets related to the US presidential candidates:

\begin{quote}
[INST] \texttt{<<SYS>>} \\
You are a helpful, respectful, and honest assistant for stance detection for presidential candidates for the USA election. Always answer from the possible options given below as helpfully as possible. Stance detection is the process of determining whether the author of a tweet is in favor of or against a given target. The target may not always be explicitly mentioned in the text, and the tweet's stance can be conveyed implicitly through subtext, regional and cultural references, or other implicit meanings. \\
The possible stances are:
\begin{itemize}
    \item support: The tweet has a positive or supportive attitude towards the target, either explicitly or implicitly.
    \item against: The tweet opposes or criticizes the target, either explicitly or implicitly.
\end{itemize}
\texttt{</SYS>} \\
Tweet: \texttt{[tweet]} \\
Stance towards the target [target]:[/INST]
\end{quote}

The placeholders \texttt{[tweet]} and \texttt{[target]} are used in a similar manner as explained for the SemEval-2016 template above.

\subsubsection{Twitter Stance 2020 LlaMa 2 Template}

This prompt template is specifically designed for analyzing tweets related to the US presidential candidates:

\begin{quote}
[INST] \texttt{<<SYS>>} \\
You are a helpful, respectful, and honest assistant for stance detection for presidential candidates for the USA election. Always answer from the possible options given below as helpfully as possible. Stance detection is the process of determining whether the author of a tweet is in favor of or against a given target. The target may not always be explicitly mentioned in the text, and the tweet's stance can be conveyed implicitly through subtext, regional and cultural references, or other implicit meanings. \\
The possible stances are:
\begin{itemize}
    \item support: The tweet has a positive or supportive attitude towards the target, either explicitly or implicitly.
    \item against: The tweet opposes or criticizes the target, either explicitly or implicitly.
    \item none: The tweet is neutral or doesn't have a stance towards the target.
\end{itemize}
\texttt{</SYS>} \\
Tweet: \texttt{[tweet]} \\
Stance towards the target [target]:[/INST]
\end{quote}

The placeholders \texttt{[tweet]} and \texttt{[target]} are used in a similar manner as explained for the SemEval-2016 template above.

\paragraph{Note on Terminology:} In the Llama 2 and Mistral templates, we decided to use the term "support" instead of "favor". This decision was made based on token analysis for Llama 2 and Mistral, revealing that the model had a specific token for "support" but not for "favor". As a result, for the sake of efficiency, "support" was used in our prompt.

\subsection{Mistral Fine-Tuning Prompts}

For the Mistral model, we utilized a prompting approach analogous to that of LLaMa-2, but with adjustments aligned to the guidelines specific for Mistral. A key difference in our implementation for LLaMa-2 was the introduction of the \texttt{<<SYS>>} marker across all prompts to signal the system's response. For Mistral, it was not required for initiating a response from the model. This uniform adaptation was applied to each dataset we analyzed. An illustrative example of a prompt used is as follows:

\subsubsection{P-Stance Mistral Template}

This prompt template is specifically designed for analyzing tweets related to the US presidential candidates:

\begin{quote}
[INST] \\
You are a helpful, respectful, and honest assistant for stance detection for a given target. Always answer from the possible options given below as helpfully as possible. Stance detection is the process of determining whether the author of a tweet is in support of or against a given target. The target may not always be explicitly mentioned in the text, and the tweet's stance can be conveyed implicitly through subtext, regional and cultural references, or other implicit meanings. \\
The possible stances are:
\begin{itemize}
    \item support: The tweet has a positive or supportive attitude towards the target, either explicitly or implicitly.
    \item against: The tweet opposes or criticizes the target, either explicitly or implicitly.
\end{itemize}
Tweet: \texttt{[tweet]} \\
Stance towards the target [target]:[/INST]
\end{quote}